\documentclass[runningheads]{llncs}
\usepackage{graphicx}
\usepackage{tikz}
\usepackage{comment}
\usepackage{amsmath,amssymb} % define this before the line numbering.
\usepackage{color}

\usepackage[accsupp]{axessibility}  % Improves PDF readability for those with disabilities.

\usepackage[width=122mm,left=12mm,paperwidth=146mm,height=193mm,top=12mm,paperheight=217mm]{geometry}
\usepackage[justification=centering]{caption}
\usepackage{hyperref}
\usepackage{academicons}
\usepackage{xcolor}

\newcommand{\orcid}[1]{\href{https://orcid.org/#1}{\includegraphics[scale=0.4]{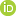}}}

\begin{document}

% \linenumbers
\pagestyle{headings}
\mainmatter
\def\ECCVSubNumber{4}  % Insert your submission number here

%\title{MC-hands-1M: The first glove-wearing hands dataset for hand pose estimation} % Replace with your title
\title{MC-hands-1M: A glove-wearing hand dataset for pose estimation} % Replace with your title

%\begin{comment}
%\titlerunning{HANDS@ECCV2022 Workshop submission ID \ECCVSubNumber} 
%\authorrunning{HANDS@ECCV2022 Workshop submission ID \ECCVSubNumber} 
\titlerunning{P. Boutis et al.} 
\authorrunning{P. Boutis et al.} 
\author{Prodromos Boutis \orcid{0000-0002-7518-1657}, Zisis Batzos \orcid{0000-0003-1595-6833}, Konstantinos Konstantoudakis \orcid{0000-0001-5092-8796}, Anastasios Dimou \orcid{0000-0003-2763-4217}, Petros Daras \orcid{0000-0003-3814-6710}}
\institute{Centre for Research \& Technology Hellas}
%\end{comment}
%******************

\begin{comment}
\titlerunning{Abbreviated paper title}
\author{First Author\inst{1}\orcidID{0000-1111-2222-3333} \and
Second Author\inst{2,3}\orcidID{1111-2222-3333-4444} \and
Third Author\inst{3}\orcidID{2222--3333-4444-5555}}
%
\authorrunning{F. Author et al.}
\institute{Princeton University, Princeton NJ 08544, USA \and
Springer Heidelberg, Tiergartenstr. 17, 69121 Heidelberg, Germany
\email{lncs@springer.com}\\
\url{http://www.springer.com/gp/computer-science/lncs} \and
ABC Institute, Rupert-Karls-University Heidelberg, Heidelberg, Germany\\
\email{\{abc,lncs\}@uni-heidelberg.de}}
\end{comment}
%******************
\maketitle

\begin{abstract}
Nowadays, the need for large amounts of carefully and complexly annotated data for the training of computer vision modules continues to grow. Furthermore, although the research community presents state of the art solutions to many problems, there exist special cases, such as the pose estimation and tracking of a glove-wearing hand, where the general approaches tend to be unable to provide an accurate solution or fail completely. In this work, we are presenting a synthetic dataset\footnote{The dataset is public and can be found at \url{https://www.zenodo.org/record/7194271/}} for 3D pose estimation of glove-wearing hands, in order to depict the value of data synthesis in computer vision. The dataset is used to fine-tune a public hand joint detection model, achieving significant performance in both synthetic and real images of glove-wearing hands.
%\footnote{The dataset will be shared after the review to maintain anonymity.}.

\end{abstract}

\section{Introduction}
Computer vision models, targeting more complex problems, are evolving at an incredible pace, resulting in an insatiable appetite for more datasets, whose size and annotations' detail are becoming a limiting factor. Therefore, in literature, the utilization of synthetic visual data, from domain adaptation techniques \cite{3} to the deployment of GANs \cite{12}, and from the Cut-Paste approach \cite{10} to video games' scenes \cite{8}, regularly combined with corresponding real data, has become an established technique over the last decade. 

Specifically, hand pose estimation is a well-studied problem with a variety of depth- \cite{cai2018weakly} and color-based \cite{29} solutions, deploying different machine-learning methods \cite{25,30}. However, many applications, in the context of hazardous work environments and sports, necessitate the use of gloves. Existing hand detection and tracking AI algorithms, trained on real \cite{moon2020interhand2,36} and synthetic \cite{ge20193d,rogez20143d} bare-hand datasets, exhibit significantly reduced performance or fail altogether in gloved hand scenarios, as they depend deeply on the canvas of the human skin's colors. Hence, there is a clear need for a gloved-hand dataset with ground truth for the joints' positions, allowing the training or re-training of AI algorithms capable of estimating poses and/or tracking hands wearing gloves of diverse size and color.

%In conclusion, the contributions of this work are: 
As a result, the contributions of this work are: 1) a synthetic image dataset for 3D pose estimation of glove-wearing hands in outdoor environments and 2) insights on its deployment to retrain hand pose estimation and tracking modules.
%\begin{itemize}
%  \item A synthetic image dataset for 3D pose estimation of glove-wearing hands in an outdoor environment.
%  \item Insights on the use of the synthetic dataset to retrain hand joint detector.
%\end{itemize}

\section{Dataset Generation}

Aiming to address the problem of 3D glove-wearing hand pose estimation, we created a Python-based, model-independent and extensible framework for the automated generation of realistic synthetic datasets, based on Blender\footnote{Open-source 3D computer graphics software toolset, \url{https://www.blender.org/}}. The adopted approach took into consideration every aspect affecting a scene's representation: background, lighting, rendering cameras and views, model's geometry, armature's poses, movement constraints and texturing related properties. Its full design and methodology will be presented in a future full-length paper.

\begin{figure}
\centering
\captionsetup{justification=centering}
\includegraphics[scale=0.3]{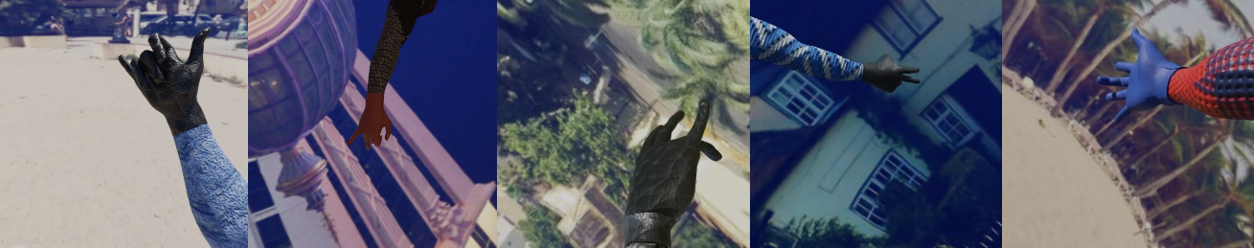}
\caption{Rendered views of the MC-hands-1M dataset}
\label{fig:mindcore_examples_5.png}
\end{figure}

%\setlength{\tabcolsep}{4pt}
%\begin{table}
%\begin{center}
%\caption{Overview of the MC-hands-1M and its different %employed variables.}
%\label{table:overview}
%\begin{tabular}{ll}
%\hline\noalign{\smallskip}
%Glove-like PBR Materials  & 10\\
%Cloth-like PBR Materials & 15 \\
%Hand's Scales & 10\\
%Wrist's (a priori) States & 3\\
%World's States (Background and Lighting) & 20\\
%Cameras & 4\\
%Views per Camera & 10\\
%\hline
%\textbf{Total Generated Images} & \textbf{1M}\\
%\hline
%\end{tabular}
%\end{center}
%\end{table}
%\setlength{\tabcolsep}{1.4pt}

As a result, we produced \emph{MC-hands-1M}, a dataset of 1M color images of a right glove-wearing hand in outdoor environments, examples of the which are depicted in Fig.\ref{fig:mindcore_examples_5.png}. The employed variables were set as follows: 10 glove- and 15 cloth-like materials, 10 hand's scales and 3 wrist's (a priori) states, 20 combinations of outdoors background with realistic sun-like lighting and 10 views per each of the 4 different cameras.

\section{Hand detection experiments and results}

The hand joint detection network selected to prove the dataset's usability is DetNet \cite{zhou2020monocular}, which receives an RGB image as input and outputs root-relative and scale-normalized 3D, as well as 2D (image space) hand joint predictions. Its architecture comprises of a feature extractor, a 2D and a 3D detector.

% Introduce testing sets (rhd and the need for ours) X
% Failure of networks on gloved test sets X
% Baseline desc X
% Experiments desc X
% Table 1 desc X
% Commenting on the results to close out X
% real+synthetic=love X
Since the existing networks trained on bare-hand datasets regularly fail to recognize glove-wearing hands, in an attempt to highlight the impact of MC-hands-1M in the alleviation of this shortcoming, we conducted a series of experiments, using test sets from the Rendered Handpose Dataset (RHD) \cite{zb2017hand} of bare and from the created MC-hands-1M of glove-wearing hands. The trained networks were compared based on the AUC-PCK metric, in order to overview the performance on images of both cases.

\setlength{\tabcolsep}{4pt}
\begin{table}
\begin{center}
\caption{Experiments' results}
\label{table:results}
\begin{tabular}{p{0.33\linewidth}  p{0.15\linewidth}  p{0.26\linewidth} }
\hline\noalign{\smallskip}
Training Sets & AUC (RHD) & AUC (MC-hands-1M)\\
\noalign{\smallskip}
\hline
\noalign{\smallskip}
RHD, CMU, GANHD & \hspace{15pt} 0.93 & \hspace{30pt} 0.19\\
MC-hands-1M & \hspace{15pt} 0.43 & \hspace{30pt} 0.97\\
MC-hands-1M, RHD, CMU & \hspace{15pt} \textbf{0.93} & \hspace{30pt} \textbf{0.97}\\ 
\hline
\end{tabular}
\end{center}
\end{table}
\setlength{\tabcolsep}{1.4pt}

Regarding the experiments, a baseline network was initially trained on the CMU Panoptic Dataset (CMU) \cite{simon2017hand}, the RHD and the GANerated Hands Dataset (GAN) \cite{29}. In the second experiment, the above baseline was retrained using images solely from the MC-hands-1M dataset. Finally, the baseline network was retrained on a mixture of real (CMU) and synthetic (RHD, MC-hands-1M) images of bare (RHD, CMU) and glove-wearing hands (MC-hands-1M).

The results of Table~\ref{table:results} depict that a network trained on traditional datasets of bare hands achieves good performance on corresponding cases, while being unable to accurately detect the joints' positions or even the existence of a hand wearing a glove. Conversely, the same network trained exclusively on synthetic glove-wearing hands' images has a significantly reduced performance on bare hands' ones. The training on a mixture of both bare and glove-wearing hand images, allows the network to achieve excellent performance for both cases. In order to examine its corresponding ability on real-life data, considering the lack of an annotated, real, non-bear hand dataset, a small collection of images exhibiting hands wearing gloves was collected.

\label{sect:experiments}
\begin{figure}
\centering
\includegraphics[width=12.55cm]{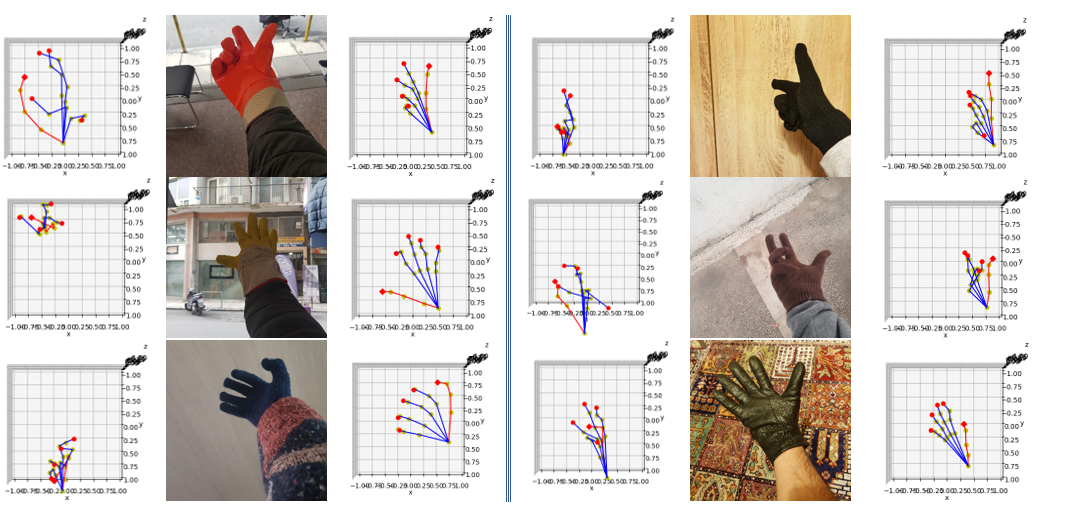}
\caption{ Examples of real images with hands wearing gloves and the corresponding outcome of the network before ({\it left}) and after ({\it right}) training on our dataset. The {\it red} lines represent the {\it index finger}}
\label{fig:before_after.png}
\end{figure}

From the visual inspection of the results, one can effortlessly observe that not only the network recognizes the hand’s existence, but it also presents a decent capability to predict correctly the orientation of the hand and the majority of the different joints' positions, as portrayed in Fig.~\ref{fig:before_after.png}.

\textbf{Funding}: This research has been supported by the European Commission within the context of the
project FASTER, funded under EU H2020 Grant Agreement 833507.

\bibliographystyle{splncs04}
\bibliography{egbib}
\end{document}